\documentclass[twoside]{article}
\usepackage{ecj,palatino,epsfig,latexsym,natbib,multirow}

\begin{document}

\ecjHeader{x}{x}{1-20}{2014}{Dynamic Swarm Dispersion in Particle Swarm Optimization for Mining Unsearched Area}{A. Bahrampour and O. Mohamad Nezami}
\title{\bf Dynamic Swarm Dispersion in Particle Swarm Optimization for Mining Unsearched Area in Solution Space (DSDPSO)}  

\author{\name{\bf Anvar Bahrampour} \hfill \addr{bahrampour@iausdj.ac.ir}\\ 
        \addr{Department of Computer Engineering, Islamic Azad University, Sanandaj Branch, Sanandaj, Iran}
 \AND
       \name{\bf Omid Mohamad Nezami} \hfill \addr{mnezami@iaubijar.ac.ir}\\
        \addr{Department of Computer Engineering, Islamic Azad University, Bijar Branch, Bijar, Iran}
}

\maketitle

\begin{abstract}

Premature convergence in particle swarm optimization (PSO) algorithm usually leads to gaining local optimum and preventing from surveying those regions of solution space which have optimal points in. In this paper, by applying special mechanisms, suitable regions were detected and then swarm was guided to them by dispersing part of particles on proper times. This process is called dynamic swarm dispersion in PSO (DSDPSO) algorithm. In order to specify the proper times and to rein the evolutionary process alternating between exploring and exploiting behaviors, we used a diversity measuring approach and implemented the dispersion mechanism. To promote the performance of DSDPSO algorithm, three different policies including particle relocation, velocity settings of dispersed particles and parameters setting were applied. We compared the promoted algorithm with similar new approaches and according to the numerical results, the proposed algorithm outperformed the basic GPSO, LPSO, DMS-PSO, CLPSO and APSO in most of the 12 standard benchmark problems with different properties taken in this study.

\end{abstract}


\section{Introduction}

The particle swarm optimization (PSO) invented by Eberhart and Kennedy (1995a,b) is applied to the concept of social interaction for problem solving \citep{31}. Each particle, denoted by $X_{i}$, represents a point in search space or a solution of the problem. The PSO algorithm iteratively modifies the point and the velocity of each particle as it searches for the optimal solution based on Equation \ref{eq:01}.\\

\qquad \qquad \qquad \quad $V_{id}=\omega*V_{id}+c_{1}r_{1}(p_{id}-x_{id})+c_{2}r_{2}(p_{gd}-x_{id})$
\begin{equation} \label{eq:01}
 X_{id}=X_{id}+V_{id}
\end{equation}
 		   
Where $V_{i}$ in the first equation is the velocity of the $i$th particle. The first part of Equation \ref{eq:01} ($ \omega*V_{id} $) is the inertia of the previous velocity in which $\omega$ is predefined by the user. The second part ($ c_{1}r_{1}(p_{id}-x_{id}) $) represents the cognition of the particle which shows personal thinking of the particle and the third part ($ c_{2}r_{2}(p_{gd}-x_{id}) $) is a social component. In this equation, $c_{1}$ and $c_{2}$ are acceleration constants. They represent weights of the stochastic acceleration terms which pull each particle toward the personal and global best positions. The constants $r_{1}$  and $r_{2}$ are the uniformly generated random numbers in interval (0, 1]. Although PSO is simple, it is a powerful search technique which has been reported to have a satisfactory performance according to many studies \citep{4}.

The convergence rate of particles in PSO is good through the fast information flowing among particles, so its diversity decreases rapidly in the successive iterations and leads to a suboptimal solution. In this case, it is said that an evolutionary process has been trapped in a local optimum or premature convergence of the process has been occured.

The standard PSO algorithm can easily get trapped in a local optimum while solving complex multimodal problems. Such deficiencies have restricted the wider applications of the PSO \citep{29,14,12}. There are several reasons why such problem arises. One of the most important reasons is decreasing diversity of the population.  A number of variants of PSO algorithm have been proposed to overcome the problem of diversity loss. One of the common methods to increase diversity is mutation. Mutation leads to the improvement of exploration abilities, which can be applied to different elements of a particle swarm. The effect of mutation depends on which elements of the swarm are mutated \citep{20}. Velocity vector mutation is equivalent to particle’s position vector mutation provided that the same mutation operator is considered.

\citet{20} proposed a negative feedback mechanism into particle swarm optimization and developed an adaptive PSO as well. This mechanism takes advantage of the swarm-diversity to control the tuning of the inertia weight (PSO-DCIW), which in turn can adjust the swarm-diversity adaptively and make a contribution to a successful global search. There are other methods including Gaussian Mutation \citep{25,8,20,23,11}, Cauchy \citep{11,24}, and Chaos Mutation \citep{5,26,28} which measure diversity and apply mutation in particles positions to improve the performance of the algorithm.

There are other ways to introduce diversity and to control its degree. \citet{29} proposed an algorithm named APSO to do so. In this method, they utilized automatic control of algorithmic parameters. A learning strategy whereby all other particles' historical best information was used to update a particle's velocity was suggested by \citet{14} and called CLPSO . \citet{19} proposed an algorithm named ARPSO in which if diversity was above the predefined threshold $d_{high}$, particles attracted each other; and if it was below $d_{low}$, particles repeled each other until they met the required high diversity $d_{high}$ . Repulsion to keep particles away from the optimum was first proposed by \citet{18}. \citet{15} made dispersion among those particles which were too close to each other; and \citet{2} reduced the attraction of the swarm centers in order to prevent the particles from tight clustering in one region of the search space and to escape from local optimum. A dynamic multi-swarm particle swarm optimizer (DMS-PSO) was proposed by \citet{13}. In this method, the whole population was divided into many small swarms. These swarms were frequently regrouped by using various regrouping schedules and information was exchanged among the swarms.

The diversity level of the swarm during the evolutionary process was measured by \citet{16}. In their study, once the diversity of the population drops down to the predefined threshold $d$, the system of generating diversified artificial particles (DAP system) is activated and starts to replace some of the particles of swarm which have relatively bad fitness by generated artificial particles (DAP particles) which have high diversity and fairly good fitness. \citet{1} and \citet{17} investigated more profoundly and promoted this basic idea by three definitions and concepts including idle particles, relocation or dispersion terms and precise search in new regions of the search space by artificial particles. They proposed a mechanism to guide the swarm based on diversity by using a diversifying process in order to detect suitable positions of the search space (points with fairly good fitness and good distance from current distribution of the swarm particles) to disperse or relocate some of the existing idle particles (those particles that there has been no change in their personal best positions for long time) in the hope of increasing diversity level of the swarm and escaping from local optimum by detecting better area of the search space. The algorithm proposed by \citet{16} was improved by defining new velocity equation for artificial particles which was used in a limited duration after each replacement to search more carefully and precisely in new regions of the search space \citep{17}. In this paper, all the previous works are engaged and a comprehensive study is conducted on the behavior of PSO algorithm with the aforementioned ideas. In addition, we prove the policies and parameters used for designing dynamic swarm dispersion particle swarm optimization (DSDPSO) algorithm.

The article is organized as follows. In section 2, diversity and measuring are defined. The DSDPSO algorithm is described in section 3. Experimental results are discussed in section 4. Finally, conclusions are mentioned in section 5.

\section{Diversity Definition and Measuring}

Population diversity is a way to monitor the degree of convergence or divergence in PSO search process  \citep{4}. There are several measures to detect diversity level of the population. Shi and Eberhart \citeyearpar{22, 21} and  \citet{30} gave three definitions for PSO population diversity measurements including position diversity, velocity diversity, and cognitive diversity. \citet{4} gave new definition for population diversity measurement, called $L1$ norm, based on both element-wise and dimension-wise diversities. They showed that useful information on search process of an optimization algorithm could be obtained by using dimension-wise definition in $L1$ norm variant. Therefore, we apply $L1$ norm of position diversity measurement in this paper. Let $m$ be the number of particles and $n$ the number of dimensions. Dimension-wise definition in $L1$ norm is defined as follows (Equation \ref{eq:02}).\\
 	
\qquad \qquad \qquad \qquad \qquad \qquad \qquad$\bar{x}=\frac{1}{m}\sum _{i=1}^{m}x_{ij}$\\

\qquad \qquad \qquad \qquad \qquad \qquad \quad $D^p = \frac{1}{m}\sum _{i=1}^{m}|x_{ij}-\bar{x_{j}}|$ 	
\begin{equation} \label{eq:02}
D^p = \frac{1}{n}\sum _{i=1}^{n}D_{j}^p
\end{equation}

Where vector $\stackrel{-}{x}$ is mean of particle position on each dimension, ${D}^{p}$ is particle position diversity vector based on $L1$ norm, and ${D}^{p}$ is the whole population diversity value.

\citet{3} introduced other approaches to measure population diversity in evolutionary computation including Hamming distance, Euclidean distance, information Entropy, and so on. In this paper, we apply Euclidean distance in particles selection process to disperse them in dispersion mechanism. The Euclidean distance, defined as Equation \ref{eq:03}, measures the distance between two particles $ X $ and $ Y $.
	
\begin{equation} \label{eq:03}
D(X,Y)=\sqrt{\sum _{i=1}^{N}(x_{i}-y_{i})^2}
\end{equation}

\section{Dynamic Swarm Dispersion Particle Swarm Optimization}

The problem of premature convergence in particle swarm optimization (PSO) algorithm often causes that the search process gets trapped in a local optimum. This problem usually occurs when the diversity of the swarm decreases and the swarm cannot escape from a local optimum. In this article, we periodically disperse some of the swarm's particles to new suitable positions with fairly good fitness in the search space and with relatively far distance from convergence point. These new points in the search space are recognized based on the history of the search process in order to enhance the diversity of the swarm and to promote the exploration ability of the algorithm. In other words, the search process should periodically  select some of the converged particles in current swarm and relocate them to new different points of the search space in order to probe new regions of the search space. Both selection process and the process of detecting new target positions of the relocated particles act based on the history of the search process up to dispersion time. In this approach, we do not change the previous personal best positions of dispersed particles in dispersion stage in order to use the result of the efforts made by relocated particles previously. Therefore, this is a relocation of the selected particles to new suitable positions rather than the replacement of them with new generated ones. Consequently, the diversity level of the swarm will be increased up to a certain degree. The evolutionary process will consistently reduce the diversity level again, and the dispersion process should be repeated periodically.

In PSO algorithm, the speed of convergence is very high, so the swarm dispersion process should be repeated frequently. On the other hand, repetition of this process is relatively time-consuming and exploitation ability of the algorithm will be decreased by high frequency of swarm dispersion, too. Therefore, we introduce a new parameter $ T $ to define the period of the dispersion system call in new search process. Another new parameter in DSDPSO algorithm which defines the amount of the swarm and should be relocated in each dispersion call is dispersion rate $ R $.

\begin{figure}[h!]
\centering
\includegraphics[width=0.9\linewidth]{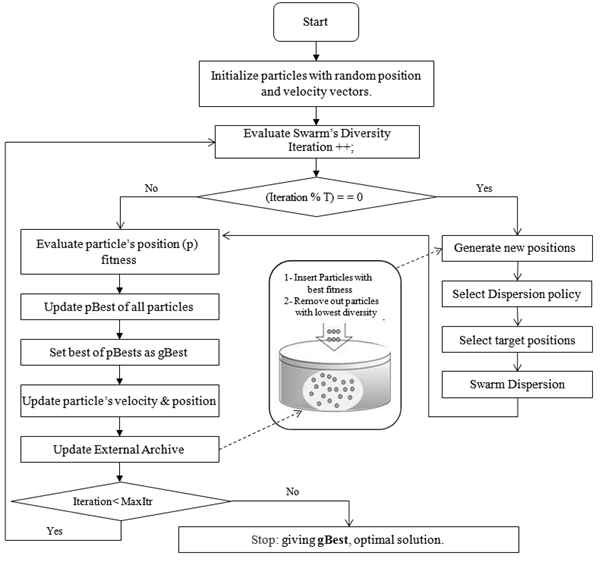}
\caption{Architecture of DSDPSO}
\label{fig:01}
\end{figure}

The architecture of a DSDPSO is illustrated in Figure \ref{fig:01}. In DSDPSO, we partially disperse swarm  to prevent from premature convergence and to extend those search spaces which are probably unexplored in the simple PSO algorithm. In Figure \ref{fig:01}, there are two different phases in process of dynamic swarm dispersion in particle swarm optimization algorithm. The first phase, which has been illustrated on the left side of Figure \ref{fig:01}, is a simple PSO plus one last step for updating an external archive to be used in dispersion mechanism. In DSDPSO, we normally execute phase one in every iteration. In the second phase, shown on the right side of Figure \ref{fig:01}, the selected particles of the swarm are dispersed to the new positions in search space in each $ T $ period. This phase will enhance the swarm diversity by relocating $R\%$ of the swarm in search space in order to escape from local optimum. 

\begin{figure}[h!]
\centering
\includegraphics[width=0.57\linewidth]{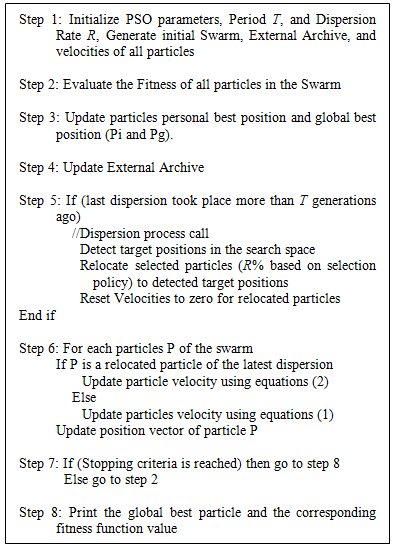}
\caption{Steps of DSDPSO algorithm}
\label{fig:02}
\end{figure}

Figure \ref{fig:02} shows the process of dynamic swarm dispersion particle swarm optimization(DSDPSO) algorithm. The steps of this process are identical to the steps of the standard PSO except for steps 5 and 6. In order to detect target positions of the selected particles for dispersion, we apply the information collected from previous generations of the PSO process using the previous global best particles to determine good points in the search space. To do so, we develop an external archive to store the global best particles of previous generations as good positions in the search space which have been visited in the hope of finding better points in the regions in which these stored particles are located. In step 4, we update the external archive if necessary; since there is no necessity for the external archive to be updated in every iteration (the process updates external archive only when notable improvement occurs in the fitness of the global best particle).

When dispersion system is called in step 5, last dispersion has taken place $ T $ generations ago. At this time, dispersion process selects $R\%$ of the swarm's particles, determines the same number of target positions, and relocates them to new positions. Dispersion process will increase the swarm diversity by relocating selected particles to new potent positions. The process of determining new target positions and selecting policy of swarm's particles for relocation will be described in the following sections. The final step of dispersion process is to reset the velocity of dispersed particles to zero because we need each dispersed particle for a very careful search in order to find better solutions in the vicinity of new location. In this paper, we found out that the previous nonzero velocities of dispersed particles are probably the cause of rapid skipping from new region and subsequently having some areas remain unsearched in the search space. This idea will be verified in section 3.2.2.

In step 6, we use two different velocity equations, Equation \ref{eq:01} and Equation \ref{eq:4} to update velocity of the swarm particles. The following velocity equation is used only for those dispersed particles generated in the latest dispersion process call. Since we want to search new regions more carefully by relocated particles, we use the minimum inertia weight. Moreover, since relocated particles are located in the regions with probably far distance from the converged particles, it is likely to have a big component of velocity to the global best particle in velocity equation. Thus, we use a random coefficient in interval (0, 1) in new velocity equation in order to relocate the particles slowly attracted to the converged swarm. This idea is verified in section 3.2.3. \\

\qquad \qquad \qquad \quad $V_{id}=r_{0}(\omega_{min}*V_{id}+c_{1}r_{1}(p_{id}-x_{id})+c_{2}r_{2}(p_{gd}-x_{id}))$
\begin{equation} \label{eq:04}
 X_{id}=X_{id}+V_{id}
\end{equation}

\begin{figure}[h!] 
\centering
\includegraphics[width=0.86\linewidth]{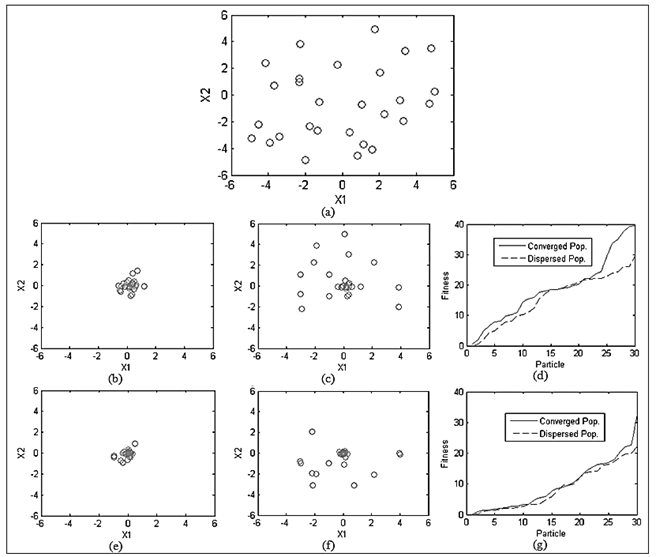}
\caption{(a) Stochastic distribution of swarm in generation 1. (b) Swarm distribution in iteration 30. (c) Swarm distribution in iteration 30 after dispersion call. (d) Fitness diagrams of swarm of b (converged pop) and c (dispersed pop) states. (e) Swarm distribution in iteration 60. (f) Swarm distribution in iteration 60 after dispersion call. (g). Fitness diagrams of swarm of e (converged pop) and f (dispersed pop) states.}
\label{fig:03}
\end{figure}

To illustrate the impact of dispersion mechanism on diversity level of the swarm in DSDPSO, we use a 2-D Rastringin function ($f_{7}$ in Table \ref{tab:01}) with the swarm of 30 particles and the dispersion rate of $45\%$. The swarm distribution in different running phases of this experiment are shown in Figure \ref{fig:03}. The stochastic initialization of particles in the first iteration is illustrated in Figure \ref{fig:03}(a). Then, the learning mechanism of the algorithm pulls particles toward the optimal region in the $ 30 $th iteration, as illustrated in Figure \ref{fig:03}(b). In the $ 30 $th iteration, DSDPSO algorithm relocates some particles of the current swarm to increase the swarm diversity based on dispersion mechanism as shown in Figure \ref{fig:03}(c). In this stage, the new swarm is named dispersed swarm. Figure \ref{fig:03}(e) and Figure \ref{fig:03}(f) show behaviors of the swarm in the $ 60 $th iteration same as Figure \ref{fig:03}(b) and Figure \ref{fig:03}(c) respectively. Figure \ref{fig:03}(c) and \ref{fig:03}(f) illustrate the achievement of dispersion mechanism for dispersing particles throughout the search space properly. It is noteworhty that in dispersion mechanism despite the increasing diversity of the swarm, the fitness level of new positions is not worse than the particles fitness before relocating. So, in DSDPSO algorithm, target positions can be generated with greater diversity and better fitness as well. Figure \ref{fig:03}(d) and \ref{fig:03}(g) substantiate this claim. 

\begin{figure}[h!]
\centering
\includegraphics[width=0.8\linewidth]{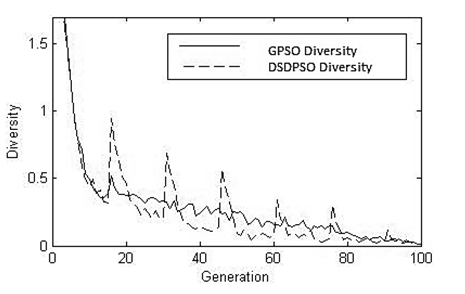}
\caption{Variation of Swarm diversity for  $f_{7}$  Function in GPSO and DSDPSO}
\label{fig:04}
\end{figure}

Finally, Figure \ref{fig:04} represents the diversity curves of standard global PSO and DSDPSO. It shows how diversity level of the swarm changes in GPSO and DSDPSO in 100 iterations. 

\newpage

\subsection{Target Positions of Selected Particles}

In this section, we describe a mechanism for determining target positions of the selected particles to disperse over the search space. In this paper, we have established an external archive with $ 100 $ particles and initialized it with random particles. Firstly, we gather those particles having the best fitness in the first $ 100 $ generations of the PSO process and use them to replace the particles in the external archive having bad fitness. Then, we should establish a replacement policy in order to gather effective particles in external archive. These particles should have both good fitness and considerable distance from the center of current distribution of the external archive particles to avoid the convergence of external archive. In this study, after the first $ 100 $ generations, replacement is applied only when the fitness of global best particle changes remarkably. In that case, one of the particles with low diversity will be removed by the new global best particle of the current swarm. To detect particles with low diversity and remove them from external archive, we use Euclidean distance described in section 2 and measure distance of each particle from the mean of the external archive particles. 

\begin{figure} [h!]
\centering
\includegraphics[width=0.9\linewidth]{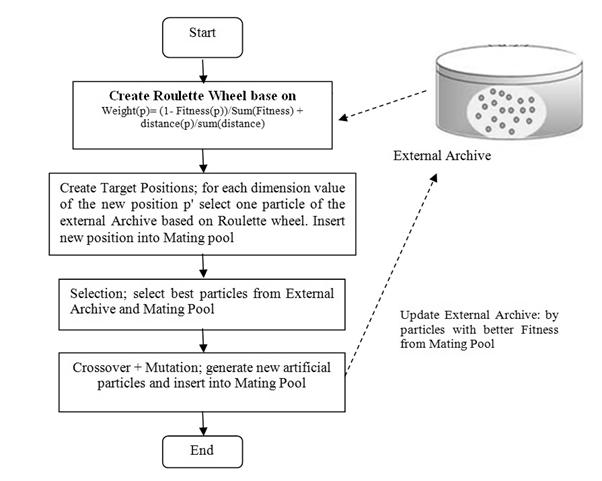}
\caption{Steps of detecting target positions in DSDPSO algorithm}
\label{fig:05}
\end{figure}

\begin{figure}[h!]
\centering
\includegraphics[width=0.6\linewidth]{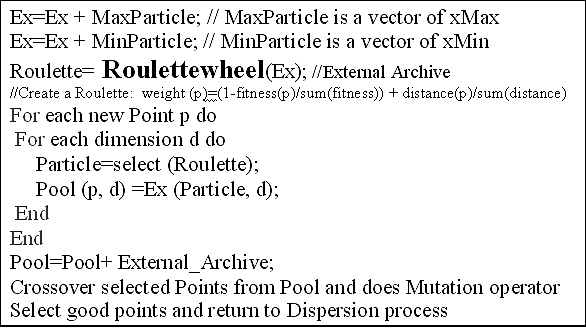}
\caption{Process of determining target positions}
\label{fig:06}
\end{figure}

Figure \ref{fig:05} illustrates the mechanism of determining target positions . To determine the new good positions for relocating converged particles of the swarm from information of the external archive, two new particles of $ x_{max} $ and $ x_{min} $ (vector of maximum or minimum value in each dimension) are added to external archive for mutation purpose. Then, a Roulette wheel is created to weigh each particle of the external archive based on its fitness and distance from the center of the swarm. In order to detect new target position in the search space, we should determine the value of each dimension of target position. Therefore, for each dimension value, one Roulette wheel selection can select a particle of the external archive, and the value stored in the same dimension of the selected particle will be used to generate the same dimension value of new target position. After selecting the value of each dimension, we  probably have a new suitable position for dispersion process. However, we do not use this point as good position for dispersion this time. We collect all the generated points ($ 100 $ points in this study) in one matting pool and add the external archive particles as good points in the search space to the pool, too. Then, operators such as genetic crossover and mutation are applied to produce new points with probably good fitness and good diversity. Afterwards, a numbers of best points ($45\%$ of the swarm in this research) are selected based on their fitness and distance from the center of the current swarm distribution and returned to dispersion process to relocate randomly the same numbers of selected particles of the swarm to these new positions. This process will remarkably increase diversity of the swarm and help to escape from local optimum trap. Figure \ref{fig:06} illustrates this process.

\subsection{Performing Policies}

As mentioned in the previous section, to promote performance of the DSDPSO algorithm, we have to properly perform several policies. In this section, we study three policies including Relocation policy, initial velocity of dispersed particles, and velocity equation of dispersed particles. These policies will be elaborated in the following three subsections. In the first subsection, we explain about different methods for selecting particles of current population for relocation and then compare them together. In the second subsection, initial velocity of dispersed particles will be explained. Finally, in the last subsection, we test different policies to control behavior of the dispersed particles and prove their abilities. In order to comparison of different approaches in each policy, we have used a collection of 10 standard benchmark problems. Mathematical models of the problems along with the true optimal value are given in Table \ref{tab:01}($f_{1}-f_{10}$). All the experiences mentioned in these subsections have been achieved under the same conditions. Because of comparing the approaches in high stability, ten standard problems with different properties were chosen to test the approaches, and average of 20 runs was applied for each one. The same initial population is used for all algorithms. The population size is specified 20, and there are 30(dimensions) for all the test problems. A linearly decreasing inertia weight starting at 0.9 and ending at 0.4 is used with the user defined parameters $c_{1}=2.0$ and $c_{2}=2.0$. For each algorithm, the maximum number of iterations is set to 3000. In all of the evaluations, a new parameter $T$ is set to 30, the external archive of size 100, and dispersion rate $R$ of 45\% are specified.

\begin{table}[h!] 
\caption{Benchmark functions used in our experimental study} 
\centering         
\begin{tabular}{p{3cm} p{6.7cm} p{2.2cm} p{0.1cm}}    
\hline\hline 
{\small Function }& {\small Function Definition} & {\small Interval} &{\small  Optimum} \\ 
\hline   
{\tiny Sphere Function}  & {\small $f_{1}(x)=\sum _{i=0}^{n-1}{x}_{i}^{2}$} & {\small [-100,100] } &{\small  0} \\ 
{\tiny  Schwefel Function 1.2} & {\small ${f}_{2}\left(x\right)=\sum _{i=0}^{n-1}{\left(\sum _{j=0}^{i}{x}_{i}\right)}^{2}$} & {\small [-100,100]} &{\small  0} \\[0.5ex]
{\tiny  Schwefel's Problem 1.2 with Noise} & {\tiny ${f}_{3}\left(x\right)=(\sum _{i=0}^{n-1}{\left(\sum _{j=0}^{i}{x}_{i}\right)}^{2})*\left(1+0.4\right(|N(0,1)|)$} &{\small  [-100,100]} & {\small 0} \\ [0.5ex]
{\tiny  Noisy Function} & {\small ${f}_{4}\left(x\right)=\left(\sum _{i=0}^{n-1}(i+1){x}_{i}^{4}\right)+rand[0,1]$} &{\small  [-1.28,1.28]} & {\small 0} \\ [0.5ex]
{\tiny  Rosenbrock Function} &{\small  ${f}_{5}\left(x\right)=\sum _{i=0}^{n-1}100{\left(x_{i+1}-x_{i}^{2}\right)}^{2}+{\left({x}_{i}-1\right)}^{2}$} & {\small [-30,30]} &{\small  0 }\\[0.5ex] 
{\tiny  Schwefel Function} & {\small ${f}_{6}\left(x\right)=418.9829n-\sum _{i= 1}^{n}{x}_{i}\sin (\sqrt{|x_{i}|})$} & {\small [-500,500]} &{\small  0} \\[0.5ex]
{\tiny  Rastrigin Function} & {\small ${f}_{7}\left(x\right)=\sum _{i= 1}^{n}(x_{i}^{2}-10\cos (2\pi x_{i})+10)$} & {\small [-5.12,5.12]} & {\small 0} \\ [0.5ex]
\multirow {3}*{\tiny  Noncaontinuous Rastrigin Function} & {\small ${f}_{8}\left(x\right)=\sum _{i= 1}^{n}[y_{i}^{2}-10\cos (2\pi y_{i})+10]$ } &{\small  [-5.12,5.12]} & {\small 0} \\ 
 & \qquad $ {y}_{i}=$ \qquad {\small  $ x_{i} \qquad \quad |x_{i}|<\frac{1}{2} , $} & & \\
  &  \qquad \qquad {\small  $\frac{round(2x_{i})}{2} \qquad|x_{i}|\ge \frac{1}{2}$} & & \\[0.5ex]
{\tiny  Shaffer’s Function} &{\small  ${f}_{9}\left(x\right)=(\sum _{i= 1}^{n}x_{i}^{2})^\frac{1}{4}[\sin ^2(50(\sum _{i= 1}^{n}x_{i}^{2})^\frac{1}{10})+1.0]$} & {\small [-32.767,32.767]} &{\small  0} \\[0.5ex] 
{\tiny  Griewank Function} & {\small ${f}_{10}\left(x\right)=\frac{1}{4000}\sum _{i= 0}^{n-1}x_{i}^{2}+\sum _{i= 0}^{n-1}\cos (\frac{x_{i}}{\sqrt{i+1}})+1$} & {\small [-600,600]} & {\small 0} \\ [0.5ex]
\multirow {2}*{\tiny  Rotated Ackley Function} &  {\small ${f}_{11}\left(x\right)=20+e-20exp(-0.2\sqrt{\frac{1}{n}\sum _{i= 1}^{n}z_{i}^{2}})-exp(\frac{1}{n}\sum _{i= 1}^{n}\cos (2\pi z_{i}))$ }& {\small [-32,32]} & {\small 0} \\ 
 & \qquad \qquad \qquad {\small $z=x*M$} & & \\[0.5ex]
\multirow {2}*{\tiny  Rotated Rastrigin's Function} & {\small ${f}_{12}\left(x\right)=\sum _{i= 1}^{n}(z_{i}^{2}-10\cos (2 \pi z_{i})+10)$} & {\small [-5.12,5.12]} & {\small 0} \\  
 & \qquad \qquad \qquad {\small $z=x*M$} & & \\[0.5ex]
\hline          
\end{tabular} 
\label{tab:01}   
\end{table} 

\subsubsection{Particles Relocation Policy}

Selecting particles and relocating them to new positions, which was completely explained in section 3.1, is performed by three approaches. In the first approach, particles with low fitness are selected for dispersion. Since probability of particles with low fitness is too low to find the optimal points of the search space, this approach is applied. It is supposed that regions around these particles are not probably suitable enough to find global optimum. In the next approach, idle particles will be relocated to new points in the search space. Idle particle is a particle that there is no change in its personal best position for long period of time; therefore, this particle could not probably find better locations.

Each of these two approaches might have some shortcomings; in the first approach, we disperse particles with low fitness, even though these particles may find better locations in subsequent generations and their fitness would be improved alternatively. In the second approach, the particles are sometimes relocated because of there is no change in their local best positions for a while, regardless to the fact that they may have high fitness. To avoid such problems, we introduce the third approach. This approach is a combination of the two previous ones. In the hybrid approach, particles with relatively low fitness and with no change in their local best positions for a long term are dispersed. Results of these approaches are indicated in Table \ref{tab:02}. According to the results, the hybrid approach leads to the best policy for dispersion.
 
\begin{table}[h!] 
\caption{The results of particles relocation approaches} 
\centering         
\begin{tabular}{c l l r}     
\hline\hline 
Relocation & Low Fitness & Idle Particle & Hybrid \\ 
\hline   
$f_{1}$ & $1.91e-4$ & $9.68e-6$ & $\mathbf{6.64e-6}$ \\ 
$f_{2}$ & $2.03e-5$ & $2.62e-7$ & $\mathbf{5.46e-8}$ \\ 
$f_{3}$ & $2.77e-2$ & $2.48e-2$ & $\mathbf{2.16e-2}$ \\ 
$f_{4}$ & $2.31e+3$ & $\mathbf{1.01e+3}$ & $1.12e+3$ \\ 
$f_{5}$  & $4.73e+1$ & $2.92e+1$ & $\mathbf{2.60e+1}$ \\ 
$f_{6}$ & $8.67e-3$ & $9.02e-4$ & $\mathbf{5.04e-4}$ \\ 
$f_{7}$ &  $6.31e-3$ & $3.95e-3$ & $\mathbf{8.62e-4}$ \\ 
$f_{8}$ & $1.23e+1$ & $4.36e+0$ & $\mathbf{3.70e+0}$ \\ 
$f_{9}$ & $1.39e+1$ & $\mathbf{7.14e+0}$ & $9.25e+0$ \\ 
$f_{10}$ & $1.42e+0$ & $1.19e+0$ & $\mathbf{1.08e+0}$ \\ [1ex]
\hline       
\end{tabular} 
\label{tab:02}   
\end{table} 

\subsubsection{Velocity Settings of Dispersed Particles}

When particles are dispersed to new positions in the search space, some of their features such as previous velocities (initial velocity after relocation), the local best positions and inertia weight are unknown. For previous velocity of a dispersed particle, we have three options including final velocity of the particle just before relocation, a random velocity, and a zero velocity. However, it is not reasonable to use previous velocity of the particle, final velocity just before relocation, in computing new velocity of the particle in new position. Random velocity causes the dispersed particles to scat from the detected new region of the search apace as the new positions are intentionally selected based on the probability of finding optimal point near new target positions. In order to avoid the latter problem and search carefully in new region, we test zero initial velocity in dispersed particles after dispersion. These three approaches are evaluated by the mentioned test functions and results are illustrated in Table \ref{tab:03}. In this experiment, the particles with low ﬁtness are selected for dispersion. According to the results, setting the initial velocities to zero in dispersion stage is the best approach.

\begin{table}[h!] 
\caption{Results of the three approaches for calculating rudimentary velocities} 
\centering         
\begin{tabular}{c l l l}     
\hline\hline 
Initial Velocity &  Random Velocity & Zero Velocity & Previous Velocity \\ 
\hline   
$f_{1}$ & $1.80e-04$ & $\mathbf{5.48e-08}$ & $1.82e-07$ \\ 
$f_{2}$ & $1.31e-05$ & $\mathbf{2.77e-11}$ & $5.07e-11$ \\ 
$f_{3}$ & $2.65e-02$ & $\mathbf{6.94e-03}$ & $8.60e-03$ \\ 
$f_{4}$ & $2.96e+03$ & $\mathbf{1.16e+03}$ & $1.23e+03$ \\ 
$f_{5}$  & $3.38e+01$ & $\mathbf{2.46e+01}$ & $2.47e+01$ \\ 
$f_{6}$ & $1.12e-02$ & $\mathbf{2.20e-06}$ & $8.07e-06$ \\ 
$f_{7}$ &  $3.30e-03$ & $\mathbf{6.16e-04}$ & $2.46e-03$ \\ 
$f_{8}$ & $1.33e+01$ & $1.62e+00$ & $\mathbf{3.81e-08}$ \\ 
$f_{9}$ & $1.32e+01$ & $2.63e+00$ & $\mathbf{1.48e+00}$ \\ 
$f_{10}$ & $1.40e+00$ & $\mathbf{2.84e-01}$ & $4.03e-01$ \\ [1ex]
\hline       
\end{tabular} 
\label{tab:03}   
\end{table} 

\subsubsection{Behavior of Dispersed Particles}

As mentioned in Section 3, since we want to search more carefully new regions of the search space by relocated particles, the dispersed particles should slowly move toward the global and local best particles. Therefore, big components of the velocity equation (Equation \ref{eq:01}) violate this aim, so we use the minimum inertia weight for dispersed particles for a period $ T $ after dispersion. Moreover, since the relocated particles are located in the regions with probably far distance from the converged particles, it is likely to have a big component of velocity to the global and local best particles in velocity equation. Thus, we use a random coefficient in interval (0, 1) in new velocity equation (Equation \ref{eq:04}) in order to relocate particles slowly attracted to the converged swarm. Equation \ref{eq:04} is applied to update velocities of the dispersed particles for $ T $ iterations after dispersion stage.

Table \ref{tab:04} shows that the new equation (Equation \ref{eq:04}) outperforms the standard velocity equation (Equation \ref{eq:01}) and low value of inertia weight in finding optimal particles. In this experiment, particles with low fitness are selected for dispersion, and the dispersed particles restart search process by initial zero velocity.

\begin{table}[h!] 
\caption{Results of the three approaches for calculating velocities} 
\centering         
\begin{tabular}{c l c l}      
\hline\hline 
Velocity &  PSO Eq. \ref{eq:01} & PSO Eq. \ref{eq:01} with Low inertia & Proposed Eq. \ref{eq:04} \\ 
\hline   
$f_{1}$ & $2.48e-04$ & $4.18e-06$ & $\mathbf{5.03e-10}$ \\ 
$f_{2}$ & $5.78e-05$ & $4.73e-10$ & $\mathbf{4.84e-15}$ \\ 
$f_{3}$ & $2.89e-02$ & $1.44e-02$ & $\mathbf{1.83e-03}$ \\ 
$f_{4}$ & $2.58e+03$ & $2.43e+03$ & $\mathbf{2.29e+03}$ \\ 
$f_{5}$  & $2.72e+01$ & $2.52e+01$ & $\mathbf{1.77e+01}$ \\ 
$f_{6}$ & $3.92e-03$ & $7.27e-05$ & $\mathbf{3.72e-09}$ \\ 
$f_{7}$ &  $3.14e-03$ & $4.06e-03$ & $\mathbf{1.13e-14}$ \\ 
$f_{8}$ & $1.13e+01$ & $\mathbf{8.27e-01}$ & $2.16e+00$ \\ 
$f_{9}$ & $1.34e+01$ & $4.19e+00$ & $\mathbf{1.51e+00}$ \\ 
$f_{10}$ & $1.57e+00$ & $1.06e+00$ & $\mathbf{3.29e-01}$ \\ 
\hline       
\end{tabular} 
\label{tab:04}   
\end{table} 

\subsection{Parameter Setting}

In order to test the effectiveness of the dispersion mechanism, a set of experimental tests were conducted to setup parameters, introduced in section 3.1, including period $T$, dispersion rate $R$, and the size of the external archive. All the experiences in this subsection have been achieved under the same conditions. To setup parameters in high stability, average of 20 runs was applied for each one. Ten standard problems with different properties were used to test the parameter values. First, the experiment was designed to setup parameter $T$. In these experiments, values of $T$ in terms of generation were set to 10, 30, 50, and 70. The final results of these experiments are shown in Table \ref{tab:05}. As we expected the lower value of $T$ led to search better fitness in the search space. Secondly, the dispersion rate $R$ was experimented by four different values of 15, 30, 45, and 60 percentage of the swarm. Table \ref{tab:06} shows the results of these experiments, the best result reached by dispersion of 45\% of swarm. Table \ref{tab:07} illustrates the results of the experiment by different external archive size. Under the condition of these experiments, the external archive with 100 particles is the best setting for this parameter.

\bigskip
\begin{table}[h!] 
\caption{Results of applying different values for $ T $ parameter} 
\centering         
\begin{tabular}{c l l l l}      
\hline\hline 
Period $ Ts $ &  10 & 30 & 50 & 70\\  
\hline  
$f_{1}$ & $4.46e-10$ & $\mathbf{4.05e-14}$ & $2.02e-13$ & $2.15e-06$ \\ 
$f_{2}$ & $1.23e-14$ & $1.45e-22$ & $\mathbf{1.11e-24}$ & $6.63e-18$ \\ 
$f_{3}$ & $\mathbf{1.31e-03}$ & $1.52e-03$ & $1.79e-03$ & $3.39e-03$ \\ 
$f_{4}$ & $\mathbf{1.95e+03}$ & $2.16e+03$ & $2.36e+03$ & $3.20e+03$ \\ 
$f_{5}$  & $\mathbf{1.96e+01}$ & $2.34e+01$ & $2.31e+01$ & $2.97e+01$ \\ 
$f_{6}$ & $2.01e-09$ & $\mathbf{1.87e-13}$ & $5.13e-13$ & $2.10e-04$ \\ 
$f_{7}$ &  $\mathbf{5.28e-03}$ & $1.23e-02$ & $1.98e-02$ & $8.35e-03$ \\ 
$f_{8}$ & $1.15e-13$ & $\mathbf{8.88e-17}$ & $6.22e-16$ & $1.14e+00$ \\ 
$f_{9}$ & $\mathbf{5.00e-02}$ & $1.17e+00$ & $6.95e+00$ & $1.38e+01$ \\ 
$f_{10}$ & $\mathbf{7.56e-02}$ & $5.04e-01$ & $6.10e-01$ & $1.06e+00$ \\ 
\hline       
\end{tabular} 
\label{tab:05}   
\end{table} 

\bigskip
  
\begin{table}[h!] 
\caption{Results of applying different dispersion rates ($ R $)} 
\centering         
\begin{tabular}{c l l l l}       
\hline\hline 
Dispersion Rate R &  15\% & 30\% & 45\% & 60\% \\  
\hline  
$f_{1}$ & $5.51e-10$ & $1.47e-13$ & $\mathbf{3.64e-14}$ & $2.59e-07$ \\ 
$f_{2}$ & $2.57e-14$ & $2.02e-21$ & $\mathbf{2.66e-22}$ & $5.25e-18$ \\ 
$f_{3}$ & $4.78e-03$ & $2.03e-03$ & $1.51e-03$ & $\mathbf{1.46e-03}$ \\ 
$f_{4}$ & $2.55e+03$ & $\mathbf{2.17e+03}$ & $2.21e+03$ & $2.50e+03$ \\ 
$f_{5}$  & $2.64e+01$ & $\mathbf{1.98e+01}$ & $2.12e+01$ & $2.22e+01$ \\ 
$f_{6}$ & $1.53e-08$ & $7.18e-13$ & $\mathbf{2.48e-13}$ & $1.54e-09$ \\ 
$f_{7}$ &  $3.20e-03$ & $\mathbf{2.71e-03}$ & $7.85e-03$ & $7.38e-03$ \\ 
$f_{8}$ & $3.08e-02$ & $\mathbf{0.00e+00}$ & $\mathbf{0.00e+00}$ & $1.47e-08$ \\ 
$f_{9}$ & $1.15e+01$ & $2.67e+00$ & $1.27e+00$ & $\mathbf{8.73e+00}$ \\ 
$f_{10}$ & $6.03e-01$ & $3.52e-01$ & $\mathbf{3.46e-01}$ & $4.83e-01$ \\ 
\hline       
\end{tabular} 
\label{tab:06}   
\end{table} 

\newpage
\begin{table}[h!] 
\caption{Results of applying different sizes of external archives} 
\centering         
\begin{tabular}{c l l l l}       
\hline\hline 
Size of External Archive &  50 & 100 & 150 & 200 \\  
\hline  
$f_{1}$ & $1.68e-10$ & $\mathbf{7.61e-14}$ & $1.70e-13$ & $5.44e-09$ \\ 
$f_{2}$ & $5.72e-18$ & $1.69e-22$ & $\mathbf{3.81e-23}$ & $2.16e-17$ \\ 
$f_{3}$ & $2.59e-03$ & $1.32e-03$ & $\mathbf{8.32e-04}$ & $1.65e-03$ \\ 
$f_{4}$ & $2.72e+03$ & $2.21e+03$ & $\mathbf{2.16e+03}$ & $2.43e+03$ \\ 
$f_{5}$  & $2.73e+01$ & $\mathbf{1.85e+01}$ & $2.21e+01$ & $2.23e+01$ \\ 
$f_{6}$ & $9.91e-11$ & $\mathbf{2.28e-13}$ & $2.59e-13$ & $8.99e-07$ \\ 
$f_{7}$ &  $8.61e-03$ & $8.47e-03$ & $\mathbf{5.04e-03}$ & $1.01e-02$ \\ 
$f_{8}$ & $4.51e-12$ & $1.78e-16$ & $\mathbf{0.00e+00}$ & $7.87e-01$ \\ 
$f_{9}$ & $5.24e+00$ & $\mathbf{2.18e+00}$ & $5.05e+00$ & $1.06e+01$ \\ 
$f_{10}$ & $7.57e-01$ & $4.25e-01$ & $3.73e-01$ & $\mathbf{2.67e-01}$ \\ 
\hline       
\end{tabular} 
\label{tab:07}   
\end{table} 

\section{Experimental Setting and Numerical Results}

In order to compare some variants of PSO and DSDPSO algorithms, we have used a collection of 12 standard benchmark problems. Mathematical models of the problems along with the true optimal value are given in TABLE 2. In this problem set, we have unimodal functions such as $f_{1}$, $f{2}$, $f{3}$ and $f{4}$.  $f{4}$ is a noisy quadric function where a uniformly distributed random variable is in the interval [0 , 1). The others are unrotated and rotated multimodal functions \citep{27}. The entire set of test problems taken for this study is scalable. In other words, the problems can be tested for any number of variables. However, in the present study, we have tested the problems for dimensions 30 and 50.	
	
\begin{figure} [h!]
\centering
\includegraphics[width=0.5\linewidth]{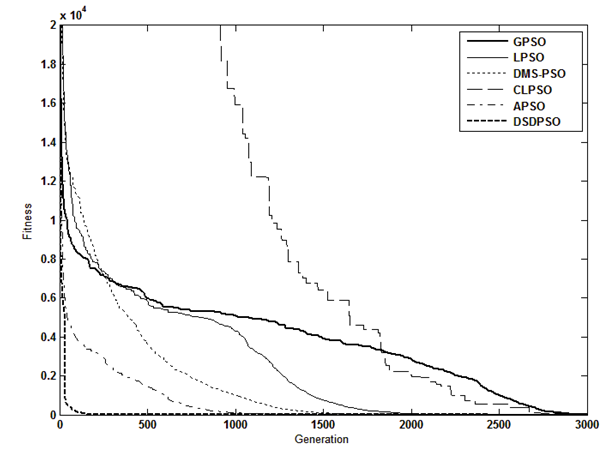}
\caption{Performance curves of GPSO, LPSO, DMS-PSO, CLPSO, APSO and DSDPSO for function $ f_{1} $}
\label{fig:07}
\end{figure}

\begin{figure} [h!]
\centering
\includegraphics[width=0.5\linewidth]{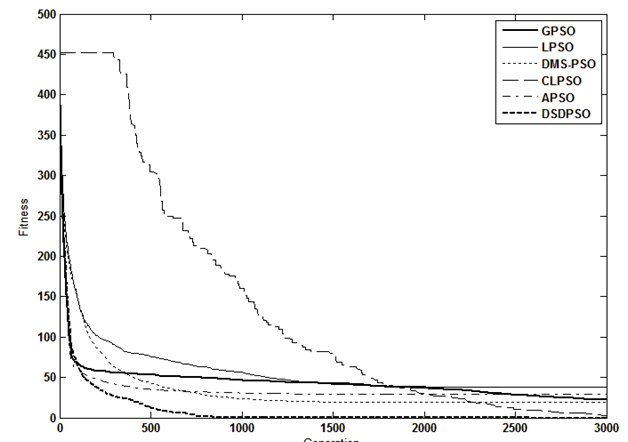}
\caption{Performance curves of GPSO, LPSO, DMS-PSO, CLPSO, APSO and DSDPSO for function $f_{5}$}
\label{fig08}
\end{figure}

\begin{figure} [h!]
\centering
\includegraphics[width=0.5\linewidth]{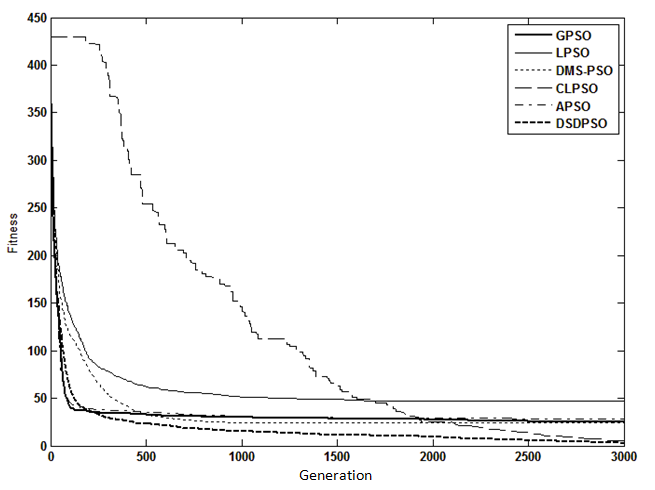}
\caption{Performance curves of GPSO, LPSO, DMS-PSO, CLPSO, APSO and DSDPSO for function $f_{8}$}
\label{fig:09}
\end{figure}

\begin{figure} [h!]
\centering
\includegraphics[width=0.5\linewidth]{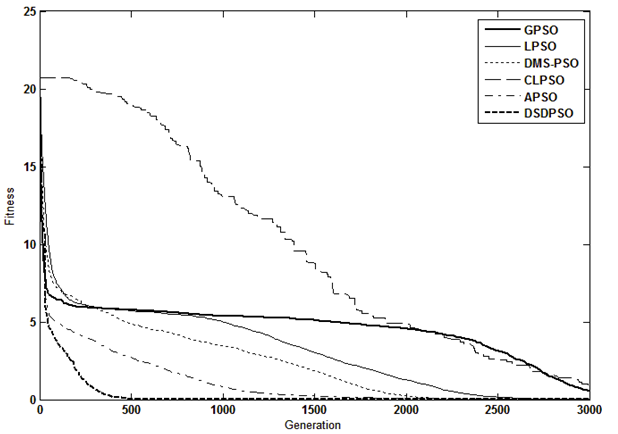}
\caption{Performance curves of GPSO, LPSO, DMS-PSO, CLPSO, APSO and DSDPSO for function $f_{11}$}
\label{fig:10}
\end{figure}

\begin{figure} [h!]
\centering
\includegraphics[width=0.5\linewidth]{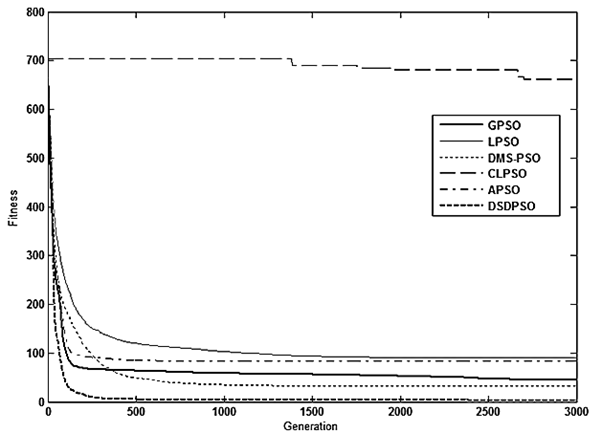}
\caption{Performance curves of GPSO, LPSO, DMS-PSO, CLPSO, APSO and DSDPSO for function $f_{12}$}
\label{fig:11}
\end{figure}

In order to make a fair comparison between the proposed DSDPSO and other variants of PSO algorithm, we implement standard PSO algorithm in both global star structure and local ring structure named GPSO and LPSO respectively. In addition, we implement APSO, CLPSO and DMS-PSO algorithms proposed in \citet{13,14,29} and compare them with DSDPSO. The same initial population is used for all algorithms. The population size is specified 20 and 50 when there are 30 and 50 variables (dimensions) respectively for all the test problems. A linearly decreasing inertia weight starting at 0.9 and ending at 0.4 is used with the user defined parameters $c_{1}=2.0$ and $c_{2}=2.0$. For each algorithm, the maximum number of iterations is set to 3000 in the case of having 30 dimensions, and 10000 for dimension 50. For DSDPSO algorithm, a new parameter $T$ is set to 30 and 50 in case of having 30 and 50 dimensions respectively. The external archive of size 100 and dispersion rate $R$ of 45\% are specified. In DMS-PSO, a group size of 3 and regroup period of 5 are applied. A total of 20 runs are conducted for each experimental setting. The results are given in Table \ref{tab:08} and Table \ref{tab:09} in terms of mean best fitness, standard deviation, and P-Value. Figure \ref{fig:07}-\ref{fig:11} show performance curves of the DSDPSO in comparison with other variants of PSO for test functions $f_{1}$, $f_{5}$, $f_{8}$, $f_{11}$, and $f_{12}$ by mean fitness of the best particles history found by the swarm in all runs. The numerical results show that the proposed algorithm outperforms other variants of PSO in most of the test cases taken in this study.

\newpage

\newpage
 
\section{Conclusion}

Evolutionary algorithms (EAs) are the best solutions for solving optimization problems. Although having different abilities to investigate the search space and attain optimal solution, they all behave similarly. One of the ideas to control the behavior of these algorithms is a rein between exploration and exploitation. In such case, we need a good mechanism to enhance the diversity of population in different stages to achieve the unsearched spaces. In order to enhance diversity and survey unsearched spaces, we used a historical approach to search and implemented that on the PSO algorithm. We proposed a mechanism to guide the swarm based on diversity by using a dispersion process in order to detect suitable positions in the search space. This model uses a dispersion mechanism to control the evolutionary process alternating between exploring and exploiting behavior and guide the algorithm, called DSDPSO algorithm, to survey the unsearched spaces. The numerical results showed that the proposed algorithm outperformed the basic GPSO, LPSO, DMS-PSO, CLPSO and APSO in most of the test cases with different properties taken in this study. It is of no doubt that this model can be used on other EAs with a little modification.

\section*{Acknowledgements}

The authors would like to thank Dr. Ponnuthurai Nagaratnam Suganthan for sending his implementations which helped us to improve the quality of our paper.

\small
\bibliographystyle{apalike}
\bibliography{w}

\newpage
\section*{APPENDIX 1 -- COMPARISON TABLES}
\begin{table}[h!] 
\caption{Comparison results GPSO, LPSO, DMS-PSO, CLPSO, APSO and DSDPSO for 20 particles of 30 dimensions in 3000 iterations} 
\centering         
\begin{tabular}{p{0.3cm} p{1.4cm} p{2cm} p{1.7cm} p{1.7cm} p{1.8cm} p{1.8cm} p{1.7cm}}      
\hline\hline 
\multicolumn {2}{c}{{\small Test Functions}} & {\small  GPSO} & {\small LPSO} & {\small DMS-PSO} & {\small CLPSO} & {\small APSO} & {\small DSDPSO} \\ 
\hline   
\multirow {3}*{$f_{1}$} &{\small  Mean} & {\small $1.7916e+01$} & {\small $5.7601e-03$} & {\small $4.6704e-06$} &{\small  $2.9291e+0$} & {\small $1.4842e-01$} & {\small $\mathbf{6.5463e-16}$}\\
& {\small Std. Dev} &{\small  $7.1447e+01$} & {\small $4.1243e-03$} & {\small $6.1429e-06$} &{\small  $9.6072e+01$} & {\small $1.7126e-01$} & {\small $\mathbf{1.3487e-15}$}\\
&{\small  P-value }& {\small $2.7607e-01$} & {\small $5.3477e-06$} & {\small $3.0037e-03$} & {\small $1.8867e-01$} & {\small $1.0176e-03$}& {\small $\mathbf{4.2828e-02}$}\\[0.5ex]

\multirow{3}*{$f_{2}$} & {\small Mean} & {\small $3.0219e+02$} & {\small $1.0972e+02$} & {\small $9.9408e+01$} & {\small $1.9861e+04$} & {\small $3.2302e+02$} &{\small  $\mathbf{1.5004e-07}$}\\ 
& {\small Std. Dev} & {\small $8.2553e+01$} &{\small  $2.9632e+0$} & {\small $4.1960e+01$} & {\small $5.8877e+03$} & {\small $8.8822e+01$} & {\small $\mathbf{1.3170e-07}$}\\
& {\small P-value} & {\small $1.1709e-12$} & {\small $9.5519e-13$} & {\small $2.0578e-09$} & {\small $4.9760e-12$ }& {\small $1.3160e-12$} & {\small $\mathbf{6.4330e-05}$}\\[0.5ex]

\multirow{3}*{$f_{3}$} &{\small  Mean} & {\small $4.3542e+02$} & {\small $3.7641e+02$} &{\small  $2.7344e+02$} & {\small $3.0351e+04$} & {\small $5.3319e+02$} & {\small $\mathbf{1.3198e-06}$}\\
& {\small Std. Dev} & {\small $8.3590e+01$} & {\small $9.8090e+01$} & {\small $1.0264e+02$} & {\small $1.7848e+04$} & {\small $7.6539e+01$} & {\small $\mathbf{1.4009e-06}$}\\
& {\small P-value} & {\small $1.9568e-15$} & {\small $5.0453e-13$} & {\small $2.9259e-10$} & {\small $3.5239e-07$} & {\small $8.9528e-18$} & {\small $\mathbf{4.7074e-04}$}\\[0.5ex]

\multirow {3}*{$f_{4}$} & {\small Mean }& {\small $5.4786e-02$} & {\small $4.7175e-02$} &{\small  $1.6212e-02$} & {\small $3.0641e-01$} & {\small $5.9757e-02$} & {\small $\mathbf{1.2017e-03}$}\\
& {\small Std. Dev} & {\small $1.2444e-02$} & {\small $9.7666e-03$} & {\small $4.9510e-03$} &{\small  $2.1417e-01$} & {\small $1.2852e-02$ }& {\small $\mathbf{4.8419e-04}$}\\
& {\small P-value} & {\small $4.2310e-14$} & {\small $7.8161e-15$} & {\small $8.3921e-12$} & {\small $3.8940e-06$} & {\small $1.5684e-14$} & {\small $\mathbf{9.5707e-10}$}\\[0.5ex]

\multirow {3}*{$f_{5}$} & {\small Mean} & {\small $3.4195e+0405$} & {\small $1.2105e+02$} & {\small $7.9335e+01$} & {\small $1.6060e+06$} & {\small $2.5087e+02$} & {\small $\mathbf{2.9116e+01}$}\\
& {\small Std. Dev} & {\small $7.1928e+04$} & {\small $5.8387e+01$} & {\small $4.9078e+01$} & {\small $3.8732e+06$} &{\small  $1.9412e+02$} & {\small $\mathbf{2.1382e+01}$}\\
&{\small  P-value }& {\small $4.6824e-02$} &{\small  $1.7544e-08$} & {\small $7.2942e-07$} & {\small $7.9286e-02$} & {\small $1.4381e-05$} & {\small $\mathbf{7.4264e-06}$}\\[0.5ex]

\multirow {3}*{$f_{6}$} & {\small Mean} & {\small $4.0397e+03$} & {\small $3.8723e+03$} & {\small $3.3921e+03$} &{\small  $\textbf{1.1476e+02}$} & {\small $1.8532e+03$} & {\small $2.7444e+03$}\\
& {\small Std. Dev} & {\small $9.7439e+02$} & {\small $3.5625e+02$} & {\small $5.9349e+02$} & {\small $\textbf{1.1291e+02}$} &{\small  $1.8849e+03$ }& {\small $4.2255e+02$}\\
& {\small P-value} & {\small $1.2563e-13$} & {\small $2.1161e-21$} & {\small $3.5365e-16$} & {\small $\textbf{2.2105e-04}$} & {\small $3.0991e-04$} & {\small $3.3025e-17$}\\[0.5ex]

\multirow {3}*{$f_{7}$} & {\small Mean} & {\small $2.2738e+01$} & {\small $3.7777e+01$} & {\small $1.8905e+01$} & {\small $3.0014e+00$} & {\small $2.8970e+01$} & {\small $\mathbf{2.6645e-16}$}\\
& {\small Std. Dev} & {\small $5.2901e+00$} & {\small $7.6694e+00$} & {\small $5.1155e+00$} & {\small $2.8917e+00$} & {\small $8.8641e+00$} & {\small $\mathbf{6.5076e-16}$}\\
&{\small  P-value} & {\small $6.5425e-14$ }& {\small $5.4632e-15$} & {\small $9.8853e-13$} & {\small $1.7781e-04$} &{\small  $8.6769e-12$ }& {\small $\mathbf{8.2814e-02}$}\\[0.5ex]

\multirow {3}*{$f_{8}$} & {\small Mean }& {\small $2.5125e+01$} & {\small $4.7025e+01$} & {\small $2.3898e+01$} &{\small  $5.2439e+00$} & {\small $2.8130e+01$} & {\small $\mathbf{2.9390e+00}$}\\
& {\small Std. Dev} & {\small $6.0741e+00$} & {\small $1.2448e+01$} & {\small $4.9217e+00$} & {\small $5.3572e+00$} & {\small $1.0252e+01$} & {\small $\mathbf{3.2009e+00}$}\\
& {\small P-value} &{\small  $1.3094e-13$} &{\small  $6.6809e-13$} & {\small $7.1008e-15$} & {\small $3.2385e-04$} & {\small $1.7760e-10$} & {\small $\mathbf{6.0109e-04}$}\\[0.5ex]

\multirow {3}*{$f_{9}$} &{\small  Mean} &{\small  $4.5303e+00$} & {\small $2.7556e+0$} & {\small $1.1980e+00$} & {\small $9.8456e+00$} & {\small $4.8676e+00$} & {\small $\mathbf{3.3620e-01}$}\\
& {\small Std. Dev} & {\small $1.1136e+00$} & {\small $1.6614e+00$} & {\small $7.7884e-01$} & {\small $4.8851e-01$} & {\small $4.2714e-01$} & {\small $\mathbf{2.7241e-01}$}\\
& {\small P-value }& {\small $1.7682e-13$} & {\small $5.0549e-07$ }& {\small $1.4605e-06$} & {\small $1.7908e-26$} & {\small $8.6800e-22$} & {\small $\mathbf{2.5255e-05}$}\\[0.5ex]

\multirow {3}*{$f_{10}$} & {\small Mean} & {\small $3.0809e-01$} & {\small $1.2467e-02$} & {\small $\textbf{1.0320e-02}$} & {\small $1.8851e+00$} & {\small $1.3385e-01$} & {\small $2.2229e-02$}\\
& {\small Std. Dev} & {\small $3.8386e-01$} & {\small $1.7777e-02$} & {\small $\textbf{1.8416e-02}$} & {\small $2.6653e+00$} & {\small $1.8158e-01$} &{\small  $2.1698e-02$}\\
& {\small P-value} & {\small $1.9550e-03$} & {\small $5.4366e-03$} & {\small $\textbf{2.1462e-02}$} & {\small $5.1211e-03$} & {\small $3.7932e-03$} & {\small $2.0376e-04$}\\[0.5ex]

\multirow {3}*{$f_{11}$} &{\small  Mean }& {\small $5.2502e-01$} & {\small $8.2130e-03$} &{\small  $1.4951e-04$} & {\small $9.1074e-01$} & {\small $1.7330e-02$ }& {\small $\mathbf{1.1623e-09}$}\\
& {\small Std. Dev} & {\small $8.8493e-01$} & {\small $8.8127e-03$} & {\small $1.8565e-04$} & {\small $1.0699e+00$ }& {\small $1.4203e-02$} & {\small $\mathbf{9.0222e-10}$}\\
& {\small P-value} & {\small $1.5690e-02$} & {\small $5.2227e-04$} & {\small $1.9023e-03$} & {\small $1.1913e-03$} & {\small $2.8975e-05$} & {\small $\mathbf{1.4964e-05}$}\\[0.5ex]

\multirow {3}*{$f_{12}$} & {\small Mean} & {\small $4.4746e+01$} & {\small $9.0639e+01$} & {\small $3.1839e+01$} & {\small $6.6216e+02$} & {\small $8.3281e+01$} & {\small $\mathbf{2.8854e+00}$}\\
&{\small  Std. Dev} & {\small $1.2894e+01$} & {\small $2.4041e+01$} & {\small $8.0443e+00$} &{\small  $8.8309e+01$} & {\small $3.1386e+01$} & {\small $\mathbf{5.8632e+00}$}\\
& {\small P-value} & {\small $3.0212e-12$} & {\small $6.9217e-13$} & {\small $2.8972e-13$} & {\small $2.2654e-18$} & {\small $3.1281e-10$} & {\small $\mathbf{4.0315e-02}$}\\[0.5ex]
\hline       
\end{tabular} 
\label{tab:08}   
\end{table} 

\newpage
\begin{table}[h!] 
\caption{Comparison results of GPSO, LPSO, DMS-PSO, CLPSO, APSO and DSDPSO for 50 particles of 50 dimensions in 10000 iterations} 
\centering               
\begin{tabular}{p{0.3cm} p{1.4cm} p{1.7cm} p{1.8cm} p{1.7cm} p{1.8cm} p{1.8cm} p{1.7cm}}     
\hline\hline 
\multicolumn {2}{c}{{\small Test Functions}} &  {\small GPSO} & {\small LPSO} & {\small DMS-PSO} & {\small CLPSO} & {\small APSO} & {\small DSDPSO} \\ 
\hline   
\multirow {3}*{$f_{1}$} & {\small Mean} & {\small $3.6468e+01$} & {\small $1.1709e-06$} & {\small $8.6047e-13$} & {\small $1.6007e-12$} & {\small $5.7967e+00$} & {\small $\mathbf{6.8476e-23}$}\\ & {\small Std. Dev} & {\small $1.2188e+02$ }& {\small $9.3600e-07$} & {\small $2.2213e-12$} &{\small  $6.0815e-12$} & {\small $2.3336e+01$} & {\small $\mathbf{2.4348e-22}$}\\
& {\small P-value} &{\small  $1.9663e-01$} & {\small $2.1459e-05$} & {\small $9.9407e-02$} & {\small $2.5371e-01$} & {\small $2.8049e-01$} & {\small $\mathbf{2.2372e-01}$}\\[0.5ex]

\multirow {3}*{$f_{2}$} & {\small Mean} & {\small $9.2720e+02$} & {\small $3.3203e+02$ }& {\small $3.4948e+02$} & {\small $3.0310e+04$} & {\small $1.0761e+03$} & {\small $\mathbf{1.9793e-09}$}\\ & {\small Std. Dev} & {\small $1.1121e+02$} & {\small $9.7868e+01$} & {\small $1.1019e+02$} & {\small $1.6111e+04$} & {\small $1.5859e+02$} & {\small $\mathbf{7.8385e-09}$}\\
& {\small P-value} & {\small $3.1056e-19$} & {\small $4.5024e-12$} & {\small $1.4669e-11$} & {\small $7.8642e-08$} & {\small $1.4614e-17$} & {\small $\mathbf{2.7284e-01}$}\\[0.5ex]

\multirow {3}*{$f_{3}$} & {\small Mean} & {\small $1.2430e+03$} &{\small  $1.4379e+03$} & {\small $7.4416e+02$} & {\small $6.2315e+04$} & {\small $1.4860e+03$} & {\small $\mathbf{3.6115e-11}$}\\
& {\small Std. Dev} & {\small $1.4253e+02$} & {\small $1.5402e+02$} & {\small $1.9543e+02$} & {\small $8.5152e+04$} & {\small $2.2121e+02$} & {\small $\mathbf{8.3015e-11}$}\\
& {\small P-value} & {\small $1.3360e-19$} & {\small $3.7150e-20$} & {\small $5.7943e-13$} & {\small $4.0035e-03$} & {\small $1.7631e-17$} & {\small $\mathbf{6.6651e-02}$}\\[0.5ex]

\multirow {3}*{$f_{4}$} &{\small  Mean} & {\small $1.4051e-01$ }& {\small $8.9698e-02$} & {\small $2.9690e-02$} & {\small $3.2467e-01$} & {\small $1.8012e-01$} & {\small $\mathbf{2.7437e-04}$}\\
& {\small Std. Dev} & {\small $2.6527e-02$ }& {\small $1.8103e-02$} & {\small $6.6752e-03$} & {\small $2.1208e-01$} & {\small $2.1536e-02$} & {\small $\mathbf{1.3640e-04}$}\\
& {\small P-value} & {\small $1.4387e-15$} & {\small $4.9015e-15$} & {\small $3.5160e-14$} & {\small $1.5604e-06$} & {\small $2.9286e-19$} & {\small $\mathbf{2.8142e-08}$}\\[0.5ex]

\multirow {3}*{$f_{5}$} & {\small Mean} & {\small $4.0587e+05$} & {\small $1.0355e+02$} & {\small $8.9578e+01$} & {\small $2.7916e+02$} & {\small $3.8231e+04$} &{\small  $\mathbf{8.2594e+00}$}\\
& {\small Std. Dev }& {\small $2.6389e+05$} & {\small $3.8853e+01$} & {\small $5.1377e+01$ }& {\small $ 4.4957e+02$} & {\small $1.4680e+05$ }& {\small $\mathbf{4.9569e+00}$}\\
& {\small P-value} & {\small $1.4632e-06$} & {\small $2.9064e-10$} &{\small  $2.4485e-07$} &{\small  $1.2011e-02$} & {\small $2.5857e-01$} & {\small $\mathbf{4.7331e-07}$}\\[0.5ex]

\multirow {3}*{$f_{6}$} &{\small  Mean} & {\small $8.1179e+03$} & {\small $6.9191e+03$} & {\small $5.9235e+03$} & {\small $6.3638e-04$ }&{\small  $7.3343e+03$} & {\small $\mathbf{4.7493e+03}$}\\
& {\small Std. Dev} & {\small $2.3530e+03$} & {\small $4.5626e+02$ }& {\small $7.5986e+02$} & {\small $6.9101e-12$} & {\small $6.0505e+03$ }& {\small $\mathbf{4.9467e+02}$}\\
& {\small P-value} & {\small $3.3496e-12$} & $3.9191e-24$ & {\small $1.0947e-18$ }& {\small $5.3075e-153$ }& {\small $3.1316e-05$} & {\small $\mathbf{2.1929e-20}$}\\[0.5ex]

\multirow {3}*{$f_{7}$} & {\small Mean} & {\small $2.8405e+01$} & {\small $5.9188e+01$} & {\small $1.8755e+01$} & {\small $4.9504e-12$} & {\small $3.1801e+01$} & {\small $\mathbf{0.0000e+00}$}\\
& {\small Std. Dev} & {\small $5.9531e+00$ }& {\small $1.1344e+01$} &{\small  $3.7126e+00$} & {\small $7.2157e-12$ }& {\small $1.3376e+01$} & {\small $\mathbf{0.0000e+00}$}\\
& {\small P-value} & {\small $9.7783e-15$} & {\small $1.8996e-15$} & {\small $3.4383e-15$} & {\small $6.3285e-03$} & {\small $1.9439e-09$} & {\small $\mathbf{NaN}$}\\[0.5ex]

\multirow {3}*{$f_{8}$} & {\small Mean} & {\small $2.9597e+01$} & {\small $6.9303e+01$} & {\small $2.3510e+01$} & {\small $4.3004e-08$} & {\small $3.5759e+01$} & {\small $\mathbf{8.8818e-17}$}\\
& {\small Std. Dev} & {\small $5.5329e+00$} & {\small $1.1018e+01$ }& {\small $6.7069e+00$} & {\small $5.2497e-08$} & {\small $1.1294e+01$} & {\small $\mathbf{3.9721e-16}$}\\
& {\small P-value} & {\small $1.2000e-15$ }& {\small $5.9973e-17$} & {\small $2.5280e-12$} & {\small $1.6516e-03$} & {\small $1.5100e-11$} & {\small $\mathbf{3.2988e-01}$}\\[0.5ex]

\multirow {3}*{$f_{9}$} & {\small Mean} & {\small $2.8477e+00$} & {\small $1.6635e+00$} &{\small  $1.2932e+00$} & {\small $1.1269e+01$} & {\small $4.7053e+00$} & $\mathbf{1.4856e-01}$\\
& Std. Dev & {\small $1.5365e+00$} & {\small $1.0521e-01$} & {\small $1.1422e-01$} & {\small $3.4260e-01$} & {\small $7.3243e-01$ }& {\small $\mathbf{2.5707e-01}$}\\
& {\small P-value} & {\small $9.8635e-08$} & {\small $1.7782e-24$} & {\small $9.7970e-22$} &{\small  $1.6480e-30$} & {\small $4.0477e-17$} & {\small $\mathbf{1.8178e-02}$}\\[0.5ex]

\multirow {3}*{$f_{10}$} & {\small Mean }& {\small $5.9085e-01$} & {\small $\mathbf{4.9700e-04}$} & {\small $1.0078e-02$ }& {\small $2.7391e-02$} & {\small $5.6336e-02$} & {\small $1.9227e-02$}\\
& {\small Std. Dev} & {\small $9.9017e-01$ }& {\small $\mathbf{2.2198e-03}$} & {\small $1.1828e-02$ }& {\small $2.1163e-02$ }& {\small $8.1117e-02$} &{\small  $2.5049e-02$}\\
& {\small P-value }& {\small $1.5182e-02$} & {\small $\mathbf{3.2928e-01}$ }&{\small  $1.1816e-03$} & {\small $1.4118e-05$} & {\small $5.8181e-03$} & {\small $2.7904e-03$}\\[0.5ex]

\multirow {3}*{$f_{11}$} &{\small  Mean} & {\small $1.2494e+00$} &{\small  $6.2859e-05$ }& {\small $1.0619e-07$} & {\small $8.9149e-09$} & {\small $1.2521e-01$} & {\small $\mathbf{1.2434e-13}$}\\
&{\small  Std. Dev} & {\small $1.8753e+00$} & {\small $1.8773e-05$ }& {\small $1.6659e-07$} & {\small $8.4447e-09$} & {\small $4.5512e-01$} & {\small $\mathbf{8.9774e-14}$}\\
& {\small P-value }& {\small $7.7015e-03$ }& {\small $5.6734e-12$ }& {\small $1.0230e-02$} &{\small  $1.4866e-04$ }& {\small $2.3360e-01$} & {\small $\mathbf{5.9585e-06}$}\\[0.5ex]

\multirow {3}*{$f_{12}$} & {\small Mean} &{\small  $6.0300e+01$} & {\small $1.2462e+02$} & {\small $2.6565e+01$ }& {\small $1.1046e+03$ }& {\small $8.2734e+01$} & {\small $\mathbf{5.3291e-16}$}\\
& {\small Std. Dev} & {\small $1.6712e+01$} & {\small $1.6415e+01$} & {\small $6.0228e+00$} &{\small  $1.0434e+02$} & {\small $2.3265e+01$ }& {\small $\mathbf{8.3518e-16}$}\\
&{\small  P-value }& {\small $1.5131e-12$} & {\small $1.7970e-18$} & {\small $4.0924e-14$ }& {\small $3.4779e-21$} & {\small $1.9582e-12$} & {\small $\mathbf{1.0163e-02}$}\\[0.5ex]
\hline       
\end{tabular} 
\label{tab:09}   
\end{table} 

\end{document}